\documentclass{article}

\usepackage{arxiv}

\usepackage[utf8]{inputenc} 
\usepackage[T1]{fontenc}    
\usepackage{hyperref}       
\usepackage{url}            
\usepackage{booktabs}       
\usepackage{amsfonts}       
\usepackage{nicefrac}       
\usepackage{microtype}      
\usepackage{lipsum}		
\usepackage{graphicx}
\usepackage[sort,numbers]{natbib}
\usepackage{doi}

\usepackage{svg}
\graphicspath{{./img/}} 
\usepackage{amssymb} 
\usepackage{amsmath}
\usepackage{subfigure}
\usepackage{algorithm}
\usepackage{algpseudocode}

\title{Linear Time Evidence Accumulation Clustering with KMeans}
\author{
	\href{https://orcid.org/0000-0002-9072-1535}{\includegraphics[scale=0.06]{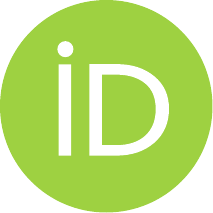}\hspace{1mm}Ga\"elle Candel} \\
	Ingenico Innovation Labs, Suresnes \\
	\texttt{firstname.lastname@ingenico.com} \\
}

\date{} 					

\renewcommand{\headeright}{}
\renewcommand{\undertitle}{}

\renewcommand{\shorttitle}{Linear Time EAC with KMeans}

\hypersetup{
pdftitle={Linear Time Evidence Accumulation Clustering with KMeans},
pdfsubject={cs.AI},
pdfauthor={Ga\"elle Candel},
pdfkeywords={Ensemble learning, Clustering, NMI, EAC},
}

\begin{document}
\maketitle

\begin{abstract}
	Among ensemble clustering methods, Evidence Accumulation Clustering is
	one of the simplest technics. In this approach, a co-association (CA)
	matrix representing the co-clustering frequency is built and then
	clustered to extract consensus clusters. Compared to other approaches,
	this one is simple as there is no need to find matches between clusters
	obtained from two different partitionings. Nevertheless, this method
	suffers from computational issues, as it requires to compute and store a
	matrix of size \(n \times n\), where \(n\) is the number of items. Due
	to the quadratic cost, this approach is reserved for small datasets.

	This work describes a trick which mimic the behavior of average linkage
	clustering. We found a way of computing efficiently the density of a
	partitioning, reducing the cost from a quadratic to linear complexity.
	Additionally, we proved that the \(k\)-means maximizes naturally the
	density. We performed experiments on several benchmark datasets where we
	compared the \(k\)-means and the bisecting version to other
	state-of-the-art consensus algorithms. The \(k\)-means' results are
	comparable to the best state of the art in terms of NMI while keeping
	the computational cost low. Additionally, the \(k\)-means led to the
	best results in terms of density. These results provide evidence that
	consensus clustering can be solved with simple algorithms.

\end{abstract}


\hypertarget{introduction}{%
\section{Introduction}\label{introduction}}

Ensemble Learning is the idea that combining multiple machine learning
models leads to a better outcome compared to a single model. The fusion
between models is done at the decision level where the final decision is
obtained thanks to a voter model.

This field started its journey with the notion of
\emph{bagging},\textsuperscript{\protect\hyperlink{ref-Breiman2004BaggingP}{1}}
where all outcomes are combined without \emph{a priori} on the models
and was democratized with
\emph{boosting},\textsuperscript{\protect\hyperlink{ref-Freund1999ASI}{2}}
where models are trained sequentially and weighted according to the
improvement they add to the group. This latest technique leads to high
accuracy improvements with much fewer models compared to bagging.

The idea of ensemble learning has multiple positive side effects. First,
model training can be speeded up, as less accurate models can lead to
better accuracy than a single large model. Next, even if models in the
ensemble belong to the same family, the decision boundaries are more
flexible. For instance, a decision tree has sharp strait boundaries,
while in an ensemble, boundaries become smooth and better adapt to data.
Another advantage is resilience, as if one model cannot run (because the
server hosting it is down, corrupted memory, network issues, invalid
input, retraining, \ldots), the inference can still be done with the
rest of the ensemble.

The field of ensemble learning is mostly known for classification, where
bagging and boosting dominate the game. There is another branch which is
about clustering, following the same idea that combining weak clustering
results together would lead to a stronger clustering. In the machine
learning area, clustering is understudied compared to classification or
regression. Clustering is mostly used for data analysis, to find
regularities within the data, to identify and extract information in a
prospective way. Because clustering is mostly about exploration, with no
direct actionable results, interest for this field is still limited.
Nevertheless, there is a lot of room for improvement and discovery.

Ensemble clustering (EClus) differs a lot from ensemble classification
(EClass), mainly in the way results are produced. In EClass, we can
identify two phases: the training phase and the inference phase. During
the training phase, each weak model is trained, possibly on its own
private dataset, and the decision function of the ensemble is set.
During the inference phase, the ensemble outputs a prediction
\(\hat{y}\) for each input \(\mathbf{x}\). The inference phase has an
unlimited lifetime, as once the ensemble is created, an unlimited number
of classifications can be done.

Ensemble clustering is also a two-step process, but the outcome of each
phase is different. In the first phase, each clustering model produces a
``weak'' partitioning of the \emph{same} dataset. In the second phase, a
categorical clustering algorithm, called the consensus algorithm,
creates a new ``strong'' partitioning using the weak partitionings.
Therefore, the lifetime of the ensemble is very short, as the consensus
algorithm serves only once.

While there are two main methods for ensemble classification, there is
no dominating strategies for ensemble clustering. Instead, there is a
myriad of technics, each with their advantages and drawbacks. There are
three branches, which depend on which information is exploited: instance
based, instance-cluster based, or cluster based algorithms.

One of the dominating approaches is the evidence accumulation clustering
(EAC) which belongs to the instance based branch. This algorithm works
in two successive phases:

\begin{itemize}

\item
  The first phase measures the co-clustering frequency, i.e., how many
  times two elements are clustered together. At the end of this step, we
  end up with a square matrix of size \(n \times n\) with values between
  \(0\) and \(1\);
\item
  The second phase is the decomposition phase, where the square matrix
  is segmented into blocks representing clusters.
\end{itemize}

This technic was popular because it is very intuitive, and when the
final partitioning is obtained, the result is visual as we should
observe blocks of \(1\) and of \(0\) when displaying the matrix with
items ordered. This method has been developed in the early 21st century,
where big data was not here yet. It appears clearly that whatever the
clustering algorithm used in the second phase, the first step prevents
from scaling because of the quadratic cost. Therefore, this technic
tends to be replaced by less memory-intensive ones.

In this article, we propose to tackle the scalability problem of this
technic. Still, it would be impossible to compute a square matrix in
linear time. However, by playing on the data representation, we can show
that the cluster's average co-clustering frequency can be computed in
linear time. By exploiting this trick, the computational cost can be
significantly decreased.

This article is decomposed the following way:

\begin{itemize}

\item
  First, we will present existing technics in the EClus field and
  present their characteristics;
\item
  Next, we describe how to represent EAC matrix data differently and
  details how the trick works;
\item
  Then, we prove the equivalence between \(k\)-means loss minimization
  and density maximization;
\item
  We describe the experimental protocol that is used to compare our
  algorithm to other well-known ensemble algorithms;
\item
  After that, we present and compare the results obtained with the
  different algorithms;
\item
  Last, we discuss our results and findings and conclude the article.
\end{itemize}

\textbf{Tags}: \emph{Ensemble learning, Evidence accumulation clustering
(EAC), Consensus clustering, NMI}

\hypertarget{review-of-existing-methods}{%
\section{Review of Existing Methods}\label{review-of-existing-methods}}

Ensemble clustering aims to create a strong clustering out of weak
clusterings, without having access to the initial features. In some way,
this is a categorical clustering problem, where the goal is to maximize
the similarity between input clusters and output result.

The field of ensemble clustering can be decomposed into three main
branches, enabling classifying algorithms by the way weak clusterings
are
represented,\textsuperscript{\protect\hyperlink{ref-MCLA}{3}},\textsuperscript{\protect\hyperlink{ref-Sample_stability}{4}}.\textsuperscript{\protect\hyperlink{ref-Graph_All}{5}}

\hypertarget{cluster-based-consensus}{%
\subsection{Cluster Based Consensus}\label{cluster-based-consensus}}

The first branch is the cluster-based one, where the consensus algorithm
works by clustering clusters, before finally assigning instance to the
best matching meta-cluster. A common way to solve this problem is using
graph partitioning
approaches,\textsuperscript{\protect\hyperlink{ref-MCLA}{3}},\textsuperscript{\protect\hyperlink{ref-EfficientScalable_CC}{6}}.\textsuperscript{\protect\hyperlink{ref-DICLENS}{7}}
A weighted \(p\)-partite graph (where \(p\) is the number of
partitioning) is constructed by measuring the similarity between all
pairs of clusters. Then, this graph is segmented, and instances are
assigned to their final cluster. Another kind of approach suggests
embedding each partitioning into a vector
space.\textsuperscript{\protect\hyperlink{ref-EC_embedding_vector_spaces}{8}}
With this transformation, it becomes easier to manipulate and compare
partitionings.

In most of these approaches, clusters are in a melting pot, where we
lose the information about their initial partitioning. Because some
partitionings might be of poor quality while others might be extremely
relevant, it is preferable to preserve this
information.\textsuperscript{\protect\hyperlink{ref-EC_Factor_graph}{9}}

The cluster-based approaches often scale very well with the number of
items, as the number of clusters often grows in
\(\mathcal{O}(\sqrt{n})\). The two drawbacks are the lack of
adaptability to instances, and the low scalability when the number of
partitioning increases. In practice, an ensemble is made of a dozen of
weak clustering, so this second limitation is often not considered.

\hypertarget{instance-cluster-based-consensus}{%
\subsection{Instance-Cluster Based
Consensus}\label{instance-cluster-based-consensus}}

The second branch gathers the instance-cluster based approaches, where
all information is preserved. Here, a weak partitioning is represented
by a binary matrix \(H^{(a)} = [\delta(i, C)]_{i=1:n, C\in \pi^a}\),
where rows represent items and columns clusters. The intersection
\((i,j)\) is filled with a \(1\) if item \(i\) belongs to cluster
\(C_j\), \(0\) otherwise. The \(p\) partitionings are concatenated into
a single binary matrix \(H = [H^{(1)}, ..., H^{(p)}]\), which is then
clustered. Therefore, each column of this matrix represents a cluster
and each row an item.

One of the simplest effective approaches is to perform a modified
\(k\)-means on this binary
matrix,\textsuperscript{\protect\hyperlink{ref-KCC}{10}},\textsuperscript{\protect\hyperlink{ref-SEC}{11}}.\textsuperscript{\protect\hyperlink{ref-ECC}{12}}
The Euclidian can be replaced by the Kullback-Leibler divergence to
better adapt to the binary case. Other approaches enhance the binary
matrix by exploiting clusters
similarity.\textsuperscript{\protect\hyperlink{ref-LCE}{13}}

Instance-cluster approaches are often preferred because they keep
information about instances and clusters until the end. However,
enhancement is often necessary as most algorithms treat features
separately.

\hypertarget{instance-based-consensus}{%
\subsection{Instance Based Consensus}\label{instance-based-consensus}}

The third branch groups instance based approaches. The most basic
approach is the evidence accumulation
clustering\textsuperscript{\protect\hyperlink{ref-CMCUEA05}{14}} (EAC)
where the co-clustering frequency between all pairs of instances is
recorded into a square matrix \(A\), which is then clustered. Because
this approach only considers item-item relationships, the cluster
information is lost at the start. To circumvent this limitation, several
approaches propose to consider clusters relationships by modifying
positive\textsuperscript{\protect\hyperlink{ref-LWEC}{15}} or negative
relationships.\textsuperscript{\protect\hyperlink{ref-ZCA}{16}}

Many clustering approaches and EClus approaches are about hard
partitioning, i.e., an item is either in or out a cluster, and never in
two clusters at the same time. Hard clustering is convenient, but it
loses the relevant information. Rather than creating a consensus
clustering with hard boundaries, other approaches suggest adding a
margin of uncertainty, where the core and the boundary of a cluster are
differentiated,\textsuperscript{\protect\hyperlink{ref-RECMC}{17}},\textsuperscript{\protect\hyperlink{ref-PNEAC}{18}}.\textsuperscript{\protect\hyperlink{ref-Self_paced_CE}{19}}

The main drawback of these approaches is the complexity, which is
firstly limited by the quadratic computational cost of the co-clustering
frequency matrix. One way to palliate to this problem is to perform
sampling and apply the ensemble algorithm on a smaller
subset.\textsuperscript{\protect\hyperlink{ref-Robust_trajectories_elite}{20}}
Despite the computational cost, these approaches produce very visual
results where blocks representing clusters are visible on the diagonal
of the co-clustering frequency
matrix,\textsuperscript{\protect\hyperlink{ref-CC_brain_matrix}{21}},\textsuperscript{\protect\hyperlink{ref-CC_RNA_network}{22}}.\textsuperscript{\protect\hyperlink{ref-PNEAC}{18}}

\hypertarget{comparison}{%
\subsection{Comparison}\label{comparison}}

It is often considered that Instance-Cluster approaches perform best
than the two others, because they keep all the information until the end
of the clustering
process,\textsuperscript{\protect\hyperlink{ref-Graph_All}{5}} and
because the complexity is (often) linear in the number of items and
clusters. However, while the full binary matrix should have all the
partitioning information in, clustering algorithms often fail to
identify relationships. It is very common to measure clusters'
similarity to enhance the binary
matrix,\textsuperscript{\protect\hyperlink{ref-LCE}{13}},\textsuperscript{\protect\hyperlink{ref-SEC}{11}}\textsuperscript{\protect\hyperlink{ref-LWEC}{15}},.\textsuperscript{\protect\hyperlink{ref-ZCA}{16}}
Because including the similarity information within the binary matrix
improves the results, it is questionable if instance-cluster approaches
are the best regarding all aspects.

\hypertarget{trick-description}{%
\section{Trick Description}\label{trick-description}}

\hypertarget{notations}{%
\subsection{Notations}\label{notations}}

\begin{itemize}

\item
  \(\mathbf{x}_i\): description of item \(i\) in the feature space
  (before clustering)
\item
  \(P = \{\pi_1, \pi_2, ..., \pi_p\}\): the set of partitionings
\item
  \(p\): number of partitioning
\item
  \(\pi = \{C_1, C_2, ..., C_k\}\): a partitioning made of \(k\)
  clusters
\item
  \(k_i\): number of cluster in partitioning \(i\)
\item
  \(|C|\): number of elements in cluster \(C\)
\item
  \(d = \sum_{i=1}^p k_i\): Total number of clusters in the ensemble
\item
  \(n\): number of items
\item
  \(A\): Co-association matrix of size \(n \times n\)
\item
  \(H\): item \(\times\) cluster binary matrix
\item
  \(\mathbf{h}_i\): description of item \(i\) in the cluster space (row
  of the matrix \(H\))
\end{itemize}

\hypertarget{base-partitionings}{%
\subsection{Base Partitionings}\label{base-partitionings}}

Be \(X = \{\mathbf{x}_i\}_{i=1:n}\) the set of \(n\) items in the
feature space. The outcome of a clustering leading to \(k\) clusters is
denoted \(\pi = \{C_1, C_2, ..., C_k\}\). We suppose that the clustering
algorithm leads to hard clusters, with the properties that:

\begin{itemize}

\item
  \(\bigcup_{C_i \in \pi} C_i = \{i\}_{i=1:n}\): The clustering covers
  all the \(n\) elements;
\item
  For any \(i \neq j\), \(C_i \cup C_j = \emptyset\): Clusters do not
  overlap.
\end{itemize}

An ensemble is made of \(p\) partitioning denoted
\(P = \{\pi_1, \pi_2, ..., \pi_p\}\). There are no constraints on the
clustering algorithm(s) used when building the ensemble, as the number
of clusters can be different in two distinct partitioning, or
partitioning can be obtained using different algorithms.

\hypertarget{co-association-matrix}{%
\subsection{Co-Association Matrix}\label{co-association-matrix}}

In evidence accumulation clustering (EAC), the co-association matrix
(CA) records the co-clustering frequency. This information is stored
into a symmetric matrix of size \(n \times n\), where each term is
computed as:

\begin{equation}
A(\mathbf{x}_i, \mathbf{x}_j) = A_{i, j} = \frac{1}{p} \sum_{\pi \in P} \sum_{C \in \pi} \delta(i, C) \delta(j, C)
\end{equation} where: \[
\delta(i, C) = \left\{
    \begin{array}{ll}
        1 & \mbox{ if } i \in C \\
        0 & \mbox{ otherwise}
    \end{array}
\right.
\]

\hypertarget{weights-and-density}{%
\subsection{Weights and Density}\label{weights-and-density}}

The most intuitive idea is to group items with a strong co-clustering
frequency, and to separate them from irrelevant elements with a low
frequency. This way, dense blocks should be visible near the diagonal
when displaying the CA matrix with elements sorted by clusters, while
low density blocks should be out of the
diagonal.\textsuperscript{\protect\hyperlink{ref-CMCUEA05}{14}}

Because performing an exhaustive search is intractable, consensus
algorithms apply heuristics to maximize the cluster coherency. When
using a hierarchical agglomerative algorithm, there are three main
heuristics:

\begin{itemize}

\item
  \emph{Single linkage}: If two groups have at least one strong link,
  they are merged together;
\item
  \emph{Average linkage}: If the average similarity between the two
  groups is high, they are merged;
\item
  \emph{Maximum linkage}: If the lowest similarity link is high, the two
  groups are merged.
\end{itemize}

Each heuristic has its own assumption about how a cluster looks like and
how we must search them. Single linkage merges groups if only a single
strong link is found, which might be too permissive. In contrast,
maximum linkage merges groups if no bad links are found, which is very
restrictive as a single wrong link prevent from merging. Average
linkage, instead, pays attention to the averaged differences between the
two groups, and merges those where this difference is negligible.

Following the idea of the average linkage which is the most reasonable
approach among the three strategies, we define in the following
paragraphs the notion of density, where the intra-density within a
cluster should be maximized while the inter-density between clusters
should instead be kept low.

\hypertarget{intra-density}{%
\subsubsection{Intra-Density}\label{intra-density}}

For a cluster \(C\), we can define its \emph{weight} as the sum of the
similarity pairs minus the diagonal elements (which are always equal to
\(1\)):

\begin{equation}
  W(C) = \sum_{i, j \in C, i \neq j} A_{i,j} = \sum_{i, j \in C} A_{i, j} - |C|
\end{equation}

From the weight, we can compute the group \emph{density} by dividing it
by the number of pairs:

\begin{equation}
  D(C) = \frac{W(C)}{|C| (|C| - 1)}
\end{equation}

It is important to remove diagonal terms as they would artificially
improve the density of small groups. In the case where a cluster has
only one element, its weight and density are set to \(0\).

\hypertarget{inter-density}{%
\subsubsection{Inter-Density}\label{inter-density}}

This quantity measures the average co-clustering frequency between items
from two different clusters. Considering two clusters \(C_0\) and
\(C_1\) with no items in common, the weight of the block is defined as:

\begin{equation}
  W(C_0, C_1) = \sum_{i \in C_0}\sum_{j \in C_1} A_{i,j}
\end{equation}

The inter density is obtained by normalizing the weight by the number of
pairs:

\begin{equation}
  D(C_0, C_1) = \frac{W(C_0, C_1)}{|C_0|.|C_1|}
\end{equation}

The average-linkage strategy merges clusters where the inter density is
high. This is supposed to maximize the overall cluster intra-density.

\hypertarget{weight-decomposition}{%
\subsection{Weight Decomposition}\label{weight-decomposition}}

The total weight of the matrix \(A\) is equal to
\(W = \sum_{i, j} A_{i, j} - n\). For a partitioning \(\pi\) (not
necessarily in \(P\)), this weight can be decomposed into two parts:

\begin{equation}
  W(\pi)  = \sum_{C_i \in \pi} W(C_i) + \sum_{C_i, C_j \in \pi, C_i \neq C_j} W(C_i, C_j)
\end{equation}

The left part is the sum of all \emph{intra}-weights, while the right
part is the sum of all \emph{inter}-weights. A good partitioning should
maximize the intra-weight while minimizing the inter-weight. However,
the weight evolves quadratically with the cluster size and will favor a
configuration with one large cluster and many small clusters, which is
not a desirable outcome.

We suggest using the weighted density where a cluster contribution is
linearly proportional to its size, which is defined as:

\begin{equation}\label{eq:density}
    S(\pi) = \frac{1}{n} \sum_{C \in \pi} |C| \times D(C)
\end{equation}

The weighted average provides a fair contribution of all clusters. Plus,
it leads to a score within \([0, 1]\) enahncing results'
interpretability whatever the cluster number. Nevertheless, this score
\emph{should not} be used to compare consensus partitioning with
different clusters' number, as more clusters will facilitate
concentration of high-density clusters near the diagonal.

\hypertarget{speed-improvement}{%
\subsection{Speed Improvement}\label{speed-improvement}}

Computing \(A\) is computationally expensive requiring
\(\mathcal{O}(p \times n^2)\) operations. If \(W\) or \(D\) are directly
computed from \(A\), these scores will also suffer from this
computational cost, even if subparts are exploited. In the following
paragraphs, we will demonstrate how \(W\) and \(D\) can be computed
efficiently.

\hypertarget{one-hot-encoding}{%
\subsubsection{One-Hot Encoding}\label{one-hot-encoding}}

In instance-cluster based approaches, the partitionings are represented
as a binary matrix \(H\) of size \(n \times d\), where
\(d = \sum_{\pi \in P} |\pi|\). A row of the matrix \(H\) is expressed
as:

\begin{equation}
    \mathbf{h}_i = [\delta(i, C)]_{C \in \{\bigcup_{\pi \in P} \pi\}}
\end{equation}

When the matrix \(H\) is available, a CA term is simply computed as:

\begin{equation}\label{eq:A_h}
    A_{i,j} = \frac{1}{p}\mathbf{h}_i^T \mathbf{h}_j
\end{equation}

When expressing \(A_{i,j}\) as the product of two \(\mathbf{h}\)
vectors, this does not improve the performance yet.

\hypertarget{weight-computation}{%
\subsubsection{Weight Computation}\label{weight-computation}}

Similarly, we can rewrite \(W\) using \(\mathbf{h}\) terms as:

\begin{equation}\label{eq:weight_h}
  \begin{array}{ll}

  W(C) &=\sum_{i, j \in C, i \neq j} A_{i,j}  \\
      &= \frac{1}{p} \times\left(\sum_{i,j \in C, i \neq j} \mathbf{h}_i\mathbf{h}_j^T\right)  \\
      &= \frac{1}{p} \times\left(\sum_{i,j \in C} \mathbf{h}_i\mathbf{h}_j^T\right) - |C|  \\
      \end{array}
\end{equation}

The density is trivially obtained from the weight by dividing by the
number of pairs \(|C| \times (|C| - 1)\). Here, we need
\(\mathcal{O}(p \times |C|^2)\) operations to compute \(W(C)\), so the
overall cost is still unchanged.

\hypertarget{shortcut-for-weight-computation}{%
\subsubsection{Shortcut for Weight
Computation}\label{shortcut-for-weight-computation}}

Weight computation can be speed up looking closely at the product
between \(\mathbf{h}_i\) and \(\mathbf{h}_j\). We can rewrite Eq.
\ref{eq:weight_h} by focusing on this product:

\begin{equation}
  p\times \left(W(C) + |C|\right)= \sum_{i, j \in C} \mathbf{h}_i \mathbf{h}_j^T = \sum_{i, j \in C} \sum_{f=1}^d h_{i, f} h_{j, f}
\end{equation}

We can exchange the order of the \(\sum\) operators as there is no
cross-dependencies:

\begin{equation}
  p \times \left(W(C) + |C|\right) = \sum_{i, j \in C} \mathbf{h}_i \mathbf{h}_j^T = \sum_{f=1}^d \sum_{i, j \in C}  h_{i, f} h_{j, f}
\end{equation}

Looking at the left sum \(\sum_{i,j \in C}\) and forgetting about the
first sum \(\sum_{f=1}^d\), we can remark that:

\begin{equation}\label{eq:tricks}
  \begin{array}{ll}
    \sum_{i, j \in C}  h_{i, f} h_{j, f} &= \sum_{i \in C} \sum_{j \in C}  h_{i, f} h_{j, f}\\
                                        &=  \left(\sum_{i \in C} h_{i, f}\right) \times \left(\sum_{j \in C} h_{j, f}\right) \\
                                        &= \left(\sum_{i \in C} h_{i, f}\right)^2
    \end{array}
\end{equation}

We can only separate \(h_{i, f}\) from \(h_{j, f}\) because \(H\) is a
binary matrix and only \(1 \times 1 = 1\). Under this form, we can
clearly see the number of positive pairs obtained thanks to column \(f\)
depends only on the number of positive elements in this column. By
denoting \(\alpha_f^C = \sum_{i \in C} h_{i, f}\), i.e., the number of
\(1\) over the \(f\)-est column for the cluster \(C\), the double sum
over clusters can be simplified into \((\alpha_f^C)^2\). Therefore,
injecting this result into Eq. \ref{eq:weight_h} the weight expression
can be simplified into:

\begin{equation}\label{eq:weight_cluster}
    W(C) = \frac{1}{p} \sum_{f=1}^d (\alpha_f^C)^2  - |C|
\end{equation}

And the density as:

\begin{equation}\label{eq:density_cluster}
    D(C) = \frac{\left(\sum_{f=1}^d \left(\alpha_f^C\right)^2\right)  - p \times |C|}{p \times  |C| (|C| - 1)}
\end{equation}

Thanks to the trick described in Eq. \ref{eq:tricks},
\(\sum_{i, j\in C} h_{i, f} h_{j, f}\) requires only
\(\mathcal{O}(|C|)\) operations instead of \(\mathcal{O}(|C|^2)\). The
weight and density can now be computed in linear time using Eq.
\ref{eq:weight_cluster} and Eq. \ref{eq:density_cluster} as we only need
to count the numbers of \(1\) for each feature.

\hypertarget{performance-comparison}{%
\subsection{Performance Comparison}\label{performance-comparison}}

Assuming we have \(t\) clusters of size \(|C| \approx \frac{n}{t}\), we
can compare the complexity of each approach.

In EAC, we need \(\mathcal{O}(p \times \frac{n^2}{t^2})\) operations per
cluster, representing a total \(\mathcal{O}(p \times \frac{n^2}{t})\).
In our approach, we need
\(\mathcal{O}(d \times |C|) = \mathcal{O}(d \times \frac{n}{t})\)
operations per cluster, leading to a total of
\(\mathcal{O}(d \times n)\). \(d\) and \(p\) have the same order of
magnitude and \(t \ll n\). Therefore, the proposed trick provides a
non-negligible speed-up.

\hypertarget{density-vs-k-means}{%
\section{\texorpdfstring{Density VS
\(k\)-means}{Density VS k-means}}\label{density-vs-k-means}}

Traditional EAC algorithms use an agglomerative algorithm to find the
final consensus
clustering.\textsuperscript{\protect\hyperlink{ref-CMCUEA05}{14}} Yet,
even if we were to apply our trick to fasten density computation, the
improvement will be limited by the inherent cost of the agglomerative
strategy.

In this section, we prove the equivalence between \(k\)-means loss
minimization and density maximization. Next, we compare hierarchical
strategies and provide support to divisive strategies. Last, we describe
the two selected clustering strategies.

\hypertarget{equivalence-between-density-and-k-means-loss}{%
\subsection{\texorpdfstring{Equivalence between Density and \(k\)-means
Loss}{Equivalence between Density and k-means Loss}}\label{equivalence-between-density-and-k-means-loss}}

Because all items have been clustered by \(p\) weak clusterings each,
the sum \(\sum_{f=1}^d \alpha^C_f = p \times |C|\) whatever \(C\).
Nevertheless, the density defined in Eq. \ref{eq:density_cluster} is
made of square terms \(\left(\alpha_f^C\right)^2\). To maximize the
overall density with this fixed budget, we need to maximize some
\(\alpha^C\) to get as close as possible from \(|C|\) while getting
other \(\alpha^C\) close to \(0\).

If we look at the traditional \(k\)-means algorithm, its goal is to
minimize the square distance between the items and the nearest prototype
vector. \begin{equation}\label{eq:loss_kmeans}
  \text{Loss}_{k\text{-means}}(\pi) = \sum_{C \in \pi} \sum_{i \in C} \| \pmb{\mu} - \mathbf{h}_i\|^2
\end{equation}

which does not seem to optimize \(\alpha^C\) terms. The prototype vector
of a cluster \(\pmb{\mu}_C\) is by definition the average of its
elements' vectors
\(\pmb{\mu}_C = \left[\frac{\alpha^C_f}{|C|}\right]_{f=1:d}\). Replacing
in Eq. \ref{eq:loss_kmeans}, the loss becomes: \begin{equation}
  \text{Loss}_{k\text{-means}}(\pi) = \sum_{C \in \pi} \sum_{i \in C} \sum_{f=1}^d \left(\frac{\alpha^C_f}{|C|} - h_{i, f}\right)^2
\end{equation}

Because \(\alpha^C_f\) represents the number of \(h_{i,f}\) equals to
\(1\), we can separate positive pairs from negative ones:
\begin{equation}
  \text{Loss}_{k\text{-means}}(\pi) = \sum_{C \in \pi}  \sum_{f=1}^d \left[\alpha^C_f \times \left(\frac{\alpha^C_f}{|C|} - 1\right)^2 + (|C| - \alpha^C_f) \times \left(\frac{\alpha^C_f}{|C|}\right)^2\right]
\end{equation} Replacing \(\alpha\) by \(\gamma\) where
\(\gamma^C_f = \frac{\alpha^C_f}{|C|}\), we can simplify the expression:
\[
\begin{array}{ll}
  \text{Loss}_{k\text{-means}}(\pi) &= \sum_{C \in \pi} |C| \times \sum_{f=1}^d  \gamma^C_f \times \left(1 - \gamma^C_f\right)^2 + (1 - \gamma^C_f) \times \left(\gamma^C_f\right)^2 \\
  &= \sum_{C \in \pi} |C| \times \sum_{f=1}^d  \gamma^C_f (1-\gamma^C_f) \\
  &= \sum_{C \in \pi} |C| \times \sum_{f=1}^d  \gamma^C_f -\left(\gamma^C_f\right)^2
\end{array}
\] Re-introducing \(\alpha\): \begin{equation}\label{eq:kmeans_loss}
  \begin{array}{ll}
    \text{Loss}_{k\text{-means}}(\pi) &= \sum_{C \in \pi} \sum_{f=1}^d  \alpha^C_f - \frac{\left(\alpha^C_f\right)^2}{|C|} \\
    &= \sum_{C \in \pi} \left(p \times |C| -   \sum_{f=1}^d  \frac{\left(\alpha^C_f\right)^2}{|C|}\right) \\
    &= p \times n - \sum_{C \in \pi}   \frac{\sum_{f=1}^d \left(\alpha^C_f\right)^2}{|C|}
    \end{array}
\end{equation}

If we express \(S(\pi)\) defined in Eq. \ref{eq:density} using the
\(\alpha\) terms, we obtain: \begin{equation}\label{eq:S_maximize}
  \begin{array}{ll}
    S(\pi) &= \frac{1}{n} \sum_{C \in \pi} |C| \times D(C) \\
        &= \frac{1}{n} \sum_{C \in \pi} |C| \frac{\left(\sum_{f=1}^d \left(\alpha_f^C\right)^2\right)  - p \times |C|}{p \times  |C| (|C| - 1)} \\
        &= \frac{1}{n} \sum_{C \in \pi}  \frac{\left(\sum_{f=1}^d \left(\alpha_f^C\right)^2\right)  - p \times |C|}{p \times  (|C| - 1)}\\
        &= \frac{1}{n} \sum_{C \in \pi} \left[\frac{\sum_{f=1}^d \left(\alpha_f^C\right)^2}{p \times  (|C| - 1)}  - \frac{p \times |C|}{p \times  (|C| - 1)} \right] \\
        &= \frac{1}{n} \sum_{C \in \pi} \left[\frac{\sum_{f=1}^d \left(\alpha_f^C\right)^2}{p \times  (|C| - 1)}  - \frac{|C|}{|C| - 1}\right] \\
        &\approx \frac{1}{np} \sum_{C \in \pi} \frac{\sum_{f=1}^d \left(\alpha_f^C\right)^2}{|C|}  - \frac{|\pi|}{n}
  \end{array}
\end{equation}

Because \(n\), \(p\) and the number of desired cluster \(|\pi|\) are
fixed, minimizing the \(k\)-means loss in Eq. \ref{eq:kmeans_loss} is
equivalent to maximizing the density in Eq. \ref{eq:S_maximize}.

\hypertarget{hierarchical-algorithm}{%
\subsection{Hierarchical Algorithm}\label{hierarchical-algorithm}}

Running the \(k\)-means would maximize the clustering density.
Nevertheless, we want to provide an alternative to hierarchical
algorithms. Indeed, while hierarchical algorithms are costly to run,
there might be more efficient when the cluster number is unknown, as
running \(n\) times the non-hierarchical algorithm would be more costly
than running once the hierarchical version. The output of a hierarchical
algorithm is a tree where relationships between items and clusters are
known. Plus, the tree allows studying coarse and fine scales at the same
time, which can be useful for some applications.

\hypertarget{families}{%
\subsubsection{Families}\label{families}}

There are two kinds of hierarchical algorithms:

\begin{itemize}

\item
  \textbf{Agglomerative} algorithms (bottom-up): The algorithm starts
  with \(n\) clusters with one element each. At each step, it searches
  for the best pair of clusters to merge, until one single big cluster
  is left;
\item
  \textbf{Divisive} algorithms (top-down): The algorithm starts with a
  single cluster with \(n\) items in. At each step, the current cluster
  is divided in two until \(n\) clusters of size \(1\) are obtained.
\end{itemize}

\hypertarget{step-complexity}{%
\subsubsection{Step Complexity}\label{step-complexity}}

For both families, the same number of \emph{steps} should be done to
build the full tree: from \(1\) to \(n\) clusters for divisive and from
\(n\) to \(1\) clusters for agglomerative. In both cases, there are
exactly \(n-1\) steps.

We should look at step complexity to understand the differences:

\begin{itemize}
\item
  When there are \(k\) clusters, an agglomerative algorithm needs to
  perform \(\frac{k(k-1)}{2}\) comparisons to find the best pair of
  clusters. Over the whole \(n-1\) steps, it leads to a complexity in
  between \(\mathcal{O}(n^2)\) and \(\mathcal{O}(n^3)\)
  (\(\mathcal{O}(n^2)\) when we keep in memory the pairs that have
  already been computed). Even if we consider the lowest bound, this
  cost prevents from scaling to large datasets.
\item
  For a divisive algorithm, we should look at the number of items in the
  cluster to be splitted. If a cluster is made of \(n\) elements, there
  are \(2^{n - 1} - 1\) ways to split this cluster in two parts. This
  cost is very prohibitive even for small clusters. Therefore, this
  family only works with heuristics.
\end{itemize}

\hypertarget{early-aborting}{%
\subsubsection{Early Aborting}\label{early-aborting}}

Whatever the dataset size, it is unlikely that someone would study
clusters of size \(1\), \(2\) or \(3\), as the amount of information is
relatively low. A hierarchical algorithm normally leads to a complete
tree where all relationships between elements are available.
Nevertheless, all this information is not relevant, and it might be
interesting to stop merging / aggregating clusters when the results will
never be exploited.

It is clear that a clustering should return at least two clusters to be
useful. Concerning the upper bound, a rule of thumb for the maximal
number of clusters to search for is \(\sqrt{n}\).

If we call \(k\) this maximal number of clusters, we can compare the
number of steps done in each approach:

\begin{itemize}
\item
  For agglomerative approaches, at least \(n-k\) steps are necessary to
  get clusters with enough items in. And then, some additional steps are
  necessary to reach the lowest bound. Therefore, a minimal of \(n-k-1\)
  and a maximal of \(n-2\) merge operations are needed to get something
  useful;
\item
  For divisive approaches, at max \(k-1\) steps are necessary to build
  the incomplete tree, as splitting more is of no use.
\end{itemize}

If we only consider the number of steps, divisive approaches are faster
than agglomerative ones. For large datasets, the difference is even more
pronounced as \(k\) grows in \(\sqrt{n}\). This is without considering
the cost of a single step. Nevertheless, if the split is done with a
\(k\)-means, the agglomerative algorithm would be highly competitive as
\(k\)-means is very fast to run.

\hypertarget{suitability-of-the-density}{%
\subsection{Suitability of the
Density}\label{suitability-of-the-density}}

A divisive algorithm can be speeded up using distributed computation if
after a split the two parts become independent. This property is not
mandatory, but it ensures that the cost of a step is independent of the
items outside the current cluster. We can verify that the density
satisfies this independence property by considering a partitioning
\(\pi = \{C_1, C_2, ..., C_k\}\) where cluster \(C_1\) is split in two.

The initial density before the split is:

\begin{equation}\label{eq:loss_density_1}
  \begin{array}{ll}
      S(\pi) &= \frac{1}{n} \sum_{C \in \pi} |C| \times D(C) \\
             &= \frac{1}{n}\left(|C_1| \times D(C_1) + \sum_{i=2}^k |C_i| \times D(C_i) \right)
    \end{array}
\end{equation}

When performing the split of cluster \(C_1\) into \(A\) and \(B\), we
obtain \(\pi^+ = \{A, B, C_2, ..., C_k\}\), where the density is:

\begin{equation}\label{eq:loss_density_2}
  \begin{array}{ll}
      S(\pi^+) &= \frac{1}{n} \sum_{C \in \pi^+} |C| \times D(C) \\
              &= \frac{1}{n}\left(|A|\times  D(A) + |B| \times D(B) + \sum_{i=2}^k |C_i|\times D(C_i)\right)
    \end{array}
\end{equation}

The difference between these two states \(\pi\) and \(\pi^+\) is:
\begin{equation}
  \begin{array}{ll}
    \Delta S &= S(\pi^+) - S(\pi) \\
    &= \frac{1}{n}\left(|A| \times D(A) + |B| \times (D(B)) - |C_1| \times D(C_1)\right)\\
    &= \frac{|C_1|}{n}\left(\beta D(A) + (1-\beta) D(B) - D(C_1)\right)
  \end{array}
\end{equation}

where \(\beta = \frac{|A|}{|A| + |B|} = \frac{|A|}{|C_1|}\). We can
clearly see that the scoring function based on the intra-density fulfill
the independence principle, as the density of clusters not involved in
the split is not impacted by the split.

\hypertarget{selected-approach}{%
\subsection{Selected Approach}\label{selected-approach}}

As \(k\)-Means tries to maximize the density, we suggest using two
complementary approaches for finding the consensus partitioning
\(\pi^*\):

\begin{itemize}

\item
  The simple \(k\)-Means applied on the binary matrix \(H\);
\item
  The bisecting
  \(k\)-Means,\textsuperscript{\protect\hyperlink{ref-BisectKMeans}{23}}
  which is an algorithm belonging to the hierarchical divisive family,
  where each split is done by performing a \(k\)-means where \(k=2\).
\end{itemize}

These two algorithms will be compared in the following sections to
state-of-the-art consensus algorithms.

\hypertarget{experimental-protocol}{%
\section{Experimental Protocol}\label{experimental-protocol}}

In this section, we describe the different elements of our experimental
protocol.

\hypertarget{comparative-algorithms}{%
\subsection{Comparative Algorithms}\label{comparative-algorithms}}

We propose to compare our proposed algorithms to several reference
algorithms. These algorithms have been re-implemented according to the
details provided in the related papers.

\hypertarget{meta-clustering-mcla-mcla}{%
\subsubsection{\texorpdfstring{Meta Clustering
(MCLA)\textsuperscript{\protect\hyperlink{ref-MCLA}{3}}}{Meta Clustering (MCLA)3}}\label{meta-clustering-mcla-mcla}}

This algorithm belongs to the cluster-based family and uses a graph
partitioning algorithm. First, the Jaccard similarity between clusters
is computed for all possible pairs, leading to a weighted graph. This
graph is segmented using the METIS
algorithm,\textsuperscript{\protect\hyperlink{ref-METIS}{24}} resulting
in a set of meta clusters. Finally, items are assigned to their most
related meta-cluster.

\hypertarget{cluster-based-similarity-partitioning-cspa-mcla}{%
\subsubsection{\texorpdfstring{Cluster-based Similarity Partitioning
(CSPA)\textsuperscript{\protect\hyperlink{ref-MCLA}{3}}}{Cluster-based Similarity Partitioning (CSPA)3}}\label{cluster-based-similarity-partitioning-cspa-mcla}}

This algorithm belongs to the instance-based family, where the CA
similarity matrix is decomposed using the METIS graph partitioning
algorithm.\textsuperscript{\protect\hyperlink{ref-METIS}{24}}

\hypertarget{spectral-ensemble-clustering-sec-sec}{%
\subsubsection{\texorpdfstring{Spectral Ensemble Clustering
(SEC)\textsuperscript{\protect\hyperlink{ref-SEC}{11}}}{Spectral Ensemble Clustering (SEC)11}}\label{spectral-ensemble-clustering-sec-sec}}

This algorithm belongs to the instance-cluster family, where a
normalized-cut spectral clustering is approximated by a weighted
\(k\)-Means. First, the matrix \(H\) is transformed into \(H'\) by
dividing each row by the weight or the ``relevance'' of the current
item. The weight is defined as:

\[w_i = \sum_{j=1}^n \mathbf{h}_j^T \mathbf{h}_i\]

This weight is proportional to its average co-clustering frequency. The
matrix to be clustered is:

\[H' = \left[\frac{\mathbf{h}_i}{w_i}\right]_{i=1:n}\]

And finally, a weighted \(k\)-means is run on \(H'\) using \(w_i\) as
the weights.

\hypertarget{entropy-based-consensus-clustering-ecc-ecc}{%
\subsubsection{\texorpdfstring{Entropy-based Consensus Clustering
(ECC)\textsuperscript{\protect\hyperlink{ref-ECC}{12}}}{Entropy-based Consensus Clustering (ECC)12}}\label{entropy-based-consensus-clustering-ecc-ecc}}

This clustering algorithm belongs to the instance-cluster family, where
the matrix \(H\) is directly segmented. This algorithm uses a modified
\(k\)-Means where the Euclidian distance has been replaced by the
Kullback-Leibler divergence.

\hypertarget{hierarchical-algorithms-cmcuea05}{%
\subsubsection{\texorpdfstring{Hierarchical
Algorithms\textsuperscript{\protect\hyperlink{ref-CMCUEA05}{14}}}{Hierarchical Algorithms14}}\label{hierarchical-algorithms-cmcuea05}}

A hierarchical agglomerative algorithm progressively merges the closest
clusters together until there is only one cluster left. There are three
ways to measure ``closeness'' between clusters:

\begin{itemize}

\item
  Single Linkage (SL): the minimal distance between any two points of
  the two clusters;
\item
  Average Linkage (AL): the average distance between all pairs of
  points;
\item
  Maximum Linkage (ML): the maximal distance that can be recorded
  between any pair of points.
\end{itemize}

There is no metric better than another and the outcome depends on the
clusters' shape and the noise level. Because of its complexity, this
algorithm cannot be used for large datasets.

We tested these algorithms on two CA matrices:

\begin{itemize}

\item
  Raw CA matrices (as described before, \(\frac{H H^T}{p}\));
\item
  Enhanced CA matrices\textsuperscript{\protect\hyperlink{ref-ZCA}{16}}
  where if \(i\) and \(j\) do not belong to the same cluster,
  \(\delta(i, C)\delta(j, C) = 0\) is replaced by
  \(\text{Sim}(C_i, C_j)\).
\end{itemize}

\hypertarget{datasets}{%
\subsection{Datasets}\label{datasets}}

The algorithms have been tested on several datasets of various sizes.
These datasets are available on the UCI
repository.\textsuperscript{\protect\hyperlink{ref-UCI}{25}} These
datasets were simply processed, by first discarding discrete variables
to keep continuous features only, and then a \(z\)-normalization was
performed to get equal contribution of all features.

\begin{table}[h]
  \centering
  \caption{Datasets description}
  \begin{tabular}{|l|r r|}
    \hline
		dataset name & \(n\) & \(f\) \\
		\hline
		\texttt{abalone} & 4177 & 8\\
		\texttt{advertisement} & 3279 & 1558\\
		\texttt{breastcancer} & 699 & 9\\
		\texttt{contraceptive} & 1473 & 9\\
		\texttt{drybean} & 13611 & 16\\
		\texttt{ecoli} & 336 & 7\\
		\texttt{gisette} & 7000 & 5000\\
		\texttt{glass} & 214 & 9\\
		\texttt{ionosphere} & 351 & 34\\
		\texttt{landstat} & 6435 & 36\\
		\texttt{letter\_recognition} & 20000 & 16\\
		\texttt{musk} & 7074 & 166\\
		\texttt{nursery} & 12960 & 27\\
		\texttt{pendigits} & 10992 & 16\\
		\texttt{user\_knowledge} & 403 & 5\\
		\texttt{yeast} & 1484 & 8\\
\hline
  \end{tabular}
\end{table}

\hypertarget{base-clustering-configuration}{%
\subsection{Base Clustering
Configuration}\label{base-clustering-configuration}}

We selected the \(k\)-means algorithm to generate the base
partitionings. This algorithm is widely used in literature because of
its simplicity and efficiency. To enhance ensemble diversity, we played
on the cluster number which was selected uniformly at random in
\([k_0, \min(\sqrt{n}, 100)]\), where \(k_0\) is the number of classes
(in the case of a classification dataset), and set to \(2\) in the case
of an unsupervised dataset.

\hypertarget{metrics}{%
\subsection{Metrics}\label{metrics}}

\hypertarget{nmi-compareclustering}{%
\subsubsection{\texorpdfstring{NMI\textsuperscript{\protect\hyperlink{ref-CompareClustering}{26}}}{NMI26}}\label{nmi-compareclustering}}

Normalized Mutual Information is a metric derived from information
theory. This is the ratio between the information shared by two
partitions \(I(\pi_A, \pi_B)\) with respect to the entropy
\(\mathcal{H}(.)\) of each partition:
\[NMI(\pi_A, \pi_B) = 2 \times \frac{I(\pi_A, \pi_B)}{\mathcal{H}(\pi_A) + \mathcal{H}(\pi_B)}\]

where \(\mathcal{H}(X)\) is the classical Shannon entropy:
\[\mathcal{H}(X) = - \sum_{x \in \mathcal{X}} P(X=x) \log_2 P(X=x)\]

and \(I(X, Y)\) is the mutual information defined as:
\[I(X, Y) = \sum_{x \in \mathcal{X}, y \in \mathcal{Y}} P(X=x, Y=y) \log_2 \left(\frac{P(X=x, Y=y)}{P(X=x)P(Y=y)}\right)\]

In the case of clusters and partitionings, \(P(X=x)\) is replaced by
\(P(x \in C | \pi = C) = \frac{|C|}{n}\), i.e., the probability of
taking an element \(x\) belonging to cluster \(C\), which is equal to
the proportion of elements in cluster \(C\). Similarly, the joint
probability
\(P(C_i, C_j | C_i \in \pi^A, C_j \in \pi^B) = \frac{|C_i \cap C_j|}{n}\)
is the probability of selecting an element belonging both to clusters
\(C_i\) and \(C_j\).

The NMI varies between \(0\) and \(1\), where \(1\) is reached when the
partitionings are identical (up to the labeling), and \(0\) if there is
no shared information.

\hypertarget{ensemble-nmi}{%
\subsubsection{Ensemble NMI}\label{ensemble-nmi}}

To evaluate a consensus algorithm, we need to compare the optimal
partitioning \(\pi^*\) to the weak partitionings \(\pi \in P\). While
the mutual information can be generalized to \(k\) variables (i.e.,
\(p\) partitionings), its computation is not practical. Instead, the
average NMI between the \(\pi^*\) and the ensemble is easier to compute:
\[\bar{NMI}(\pi^*) = \frac{1}{p} \sum_{i=1}^p NMI(\pi^*, \pi_i)\]

\hypertarget{best-and-mean}{%
\subsubsection{Best and Mean}\label{best-and-mean}}

We suggest looking at the mean and best ensemble NMI to get an idea of
values obtained with zero effort. They are defined as:

\begin{itemize}

\item
  Mean:
  \(\frac{1}{p (p-1)}\sum_{\pi_i, \pi_j \in P, i \neq j} M(\pi_i, \pi_j)\)
\item
  Best:
  \(\max_{\pi_i \in P} \frac{1}{p-1}\sum_{\pi_j \in P, i \neq j} M(\pi_i, \pi_j)\)
\end{itemize}

The ``mean'' corresponds to the average score we would get when
selecting at random a base clustering as the consensus one. The ``best''
corresponds to the best \(\bar{NMI}\) obtained when selecting a base
clustering as the optimal partitioning. These measures allow estimating
the dataset difficulty, where high values leave small room for
improvement. However, it is to note that these values are not NMI's
baseline or bound. Indeed, it is irrelevant to compare NMIs when the
number of clusters is different, as it would be irrelevant to compare
two compression algorithms when the resulting files are of different
size.

\hypertarget{density}{%
\subsubsection{Density}\label{density}}

Algorithm results are compared in terms of density (defined in Eq.
\ref{eq:density}). This metric is complementary to the NMI as it has a
different behavior. The NMI pays attention to matching pairs and
discording boundaries, and maximal score is reached if clusters
perfectly overlap. In contrast, the density pays only attention to
intra-cluster homogeneity, and adds no penalty if matching pairs have
been leftover. A density of \(1\) can be reached if an unlimited number
of clusters is allowed, while an NMI of \(1\) cannot be reached in an
ensemble clustering setup. We record as for the NMI the mean and best
density reachable when selecting a weak partitioning from the ensemble.

\hypertarget{evaluation-protocol}{%
\subsection{Evaluation Protocol}\label{evaluation-protocol}}

For each dataset, we performed \(20\) times the following experiment:

\begin{enumerate}
\def\labelenumi{\arabic{enumi}.}

\item
  Create \(20\) base clusterings:

  \begin{itemize}

  \item
    For each clustering, select uniformly at random
    \(k \in [k_0, \min(100, \sqrt{n})]\)
  \item
    Record the average and maximal NMI / density
  \end{itemize}
\item
  For each consensus algorithms:

  \begin{itemize}

  \item
    Apply the consensus algorithm to find \(20\) clusters per partition;
  \item
    Record the average NMI and density between the consensus
    partitioning and the ensemble.
  \end{itemize}
\end{enumerate}

The algorithm's final score is the average over the \(20\) iterations of
this loop.

We arbitrarily fixed the number of clusters to search to \(20\), as
moving from \(20\) to \(25\) (for instance) would have a limited impact
on the numerical results and would not impact the algorithm behavior nor
final ranking. Testing all cluster numbers to find the best NMI would be
very time-consuming with very small benefit on the overall results.

\hypertarget{results}{%
\section{Results}\label{results}}

For readability, the results are splitted into three groups of tables,
each with two tables (one for NMI and one for density):

\begin{itemize}

\item
  The former group records the average score of the non-hierarchical
  algorithms over all datasets;
\item
  The second set gathers results for agglomerative hierarchical
  algorithms for small datasets only;
\item
  The last tables compare the best score of each family and compare them
  to the values obtained with the simple \(k\)-means (KM) and the
  bisecting \(k\)-means (H-KM), plus the mean and best values obtained
  when selecting a partitioning from the ensemble.
\end{itemize}

\hypertarget{non-hierarchical-methods}{%
\subsection{\texorpdfstring{Non-Hierarchical Methods
\label{non-hierarchical}}{Non-Hierarchical Methods }}\label{non-hierarchical-methods}}

In the following tables are reported the NMI and density obtained with
the four non-hierarchical algorithms of the literature.

\begin{table}[h]
  \centering
  \caption{NMI for non-hierarchical algorithms}
  \begin{tabular}{|c|c|c|c|c|}
    \hline
		Dataset & SEC & MCLA & ECC & CSPA\\
		\hline
		\texttt{abalone} & 77.91 & \textbf{81.02} & 73.74 & 79.87\\
		\texttt{advertisement} & \textbf{80.61} & 79.93 & 80.03 & 67.62\\
		\texttt{breastcancer} & 70.13 & \textbf{74.87} & 72.46 & 60.91\\
		\texttt{contraceptive} & \textbf{74.39} & 73.21 & 73.23 & 70.39\\
		\texttt{drybean} & 78.5 & \textbf{78.9} & 69.99 & 78.22\\
		\texttt{ecoli} & 79.74 & \textbf{80.91} & 79.15 & 71.23\\
		\texttt{gisette} & \textbf{59.82} & 59.62 & 52.91 & 54.81\\
		\texttt{glass} & 77.39 & \textbf{84.3} & 76.86 & 61.03\\
		\texttt{ionosphere} & 71.72 & \textbf{74.14} & 69.47 & 		58.64\\
		\texttt{landstat} & 75.55 & \textbf{76.06} & 71.21 & 		70.06\\
		\texttt{musk} & 77.85 & \textbf{78.78} & 68.79 & 76.22\\
		\texttt{nursery} & 63.7 & \textbf{65.06} & 52.44 & 64.88\\
		\texttt{pendigits} & 76.13 & \textbf{79.69} & 68.97 & 		76.73\\
		\texttt{user\_knowledge} & \textbf{76.33} & 74.5 & 74.31 & 		55.84\\
		\texttt{yeast} & \textbf{72.83} & 71.67 & 70.2 & 68.39\\
	\hline

	  \end{tabular}
	\end{table}

In the NMI table, we can clearly see that the two leading algorithms are
SEC and MCLA algorithms which often share the first and second place,
with a very small difference, while ECC and CSPA have non-consistent
performances, with sometimes more than 5 points of differences with the
best value.

\begin{table}[h]
  \centering
  \caption{Density for non-hierarchical algorithms}
  \begin{tabular}{|c|c|c|c|c|}
    \hline
		Dataset & SEC & MCLA & ECC & CSPA\\
		\hline
		\texttt{abalone} & 45.5 & 41.6 & 37.7 & \textbf{47.5}\\
		\texttt{advertisement} & 60.4 & 71.6 & \textbf{77.2} & 		70.2\\
		\texttt{breastcancer} & 78.6 & 76.9 & \textbf{82.3} & 		76.3\\
		\texttt{contraceptive} & \textbf{57.9} & 55.7 & 48.5 & 		55.3\\
		\texttt{drybean} & 50.5 & 41.3 & 37.0 & ~\textbf{52.1}\\
		\texttt{ecoli} & \textbf{81.2} & 76.1 & 79.7 & 72.3\\
		\texttt{gisette} & \textbf{34.5} & 32.7 & 26.3 & 31.2\\
		\texttt{glass} & 93.1 & 87.7 & \textbf{94.6} & 84.7\\
		\texttt{ionosphere} & \textbf{86.3} & 81.0 & 85.7 & 78.9\\
		\texttt{landstat} & \textbf{51.6} & 47.2 & 43.4 & 51.1\\
		\texttt{musk} & 48.7 & 50.9 & 38.0 & \textbf{51.8}\\
		\texttt{nursery} & 27.8 & \textbf{30.8} & 18.2 & 30.3\\
		\texttt{pendigits} & 40.4 & 41.2 & 31.4 & \textbf{43.1}\\
		\texttt{user\_knowledge} & \textbf{78.0} & 71.7 & 75.1 & 		53.1\\
		\texttt{yeast} & \textbf{58.1} & 55.4 & 52.5 & 55.6\\
		\hline
	\end{tabular}
\end{table}

Concerning the density, there is no algorithm better than the others.

\hypertarget{hierarchical-methods}{%
\subsection{\texorpdfstring{Hierarchical Methods
\label{hierarchical}}{Hierarchical Methods }}\label{hierarchical-methods}}

In the following table, we report the results of hierarchical
agglomerative algorithms. We distinguish the results obtained using the
raw CA matrix (denoted by ``R-'') from the results obtained with the
enhanced matrix (denoted ``E-''). We add suffixes (-SL, -AL, -ML) to
distinguish the way aggregation is performed. Because of the memory
cost, these algorithms were not applied on datasets larger than the
\texttt{landstat} dataset.

\begin{table}[h]
  \centering
  \caption{NMI for hierarchical algorithms}
  \begin{tabular}{|c|c|c|c|c|c|c|}
    \hline

Dataset & R-SL & R-AL & R-ML & E-SL & E-AL & E-ML\\
\hline
\texttt{abalone} & 51.53 & \textbf{81.31} & 55.32 & 52.86 & 80.55 & 81.06\\
\texttt{advertisement} & 59.27 & \textbf{81.93} & 78.63 & 60.21 & 80.19 & 79.71\\
\texttt{breastcancer} & 67.29 & \textbf{77.76} & 76.86 & 65.12 & 73.2 & 74.1\\
\texttt{contraceptive} & 45.87 & \textbf{75.28} & 72.6 & 50.09 & 74.69 & 73.7\\
\texttt{ecoli} & 80.68 & \textbf{82.61} & 81.78 & 76.9 & 82.62 & 82.23\\
\texttt{glass} & 83.07 & 81.67 & 81.22 & 84.0 & \textbf{84.01} & 83.48\\
\texttt{ionosphere} & 71.94 & \textbf{76.47} & 74.18 & 73.52 & 76.19 & 75.66\\
\texttt{landstat} & 5.28 & \textbf{76.26} & 58.33 & 6.53 & 73.23 & 74.02\\
\texttt{user\_knowledge} & 69.66 & \textbf{77.89} & 77.34 & 67.27 & 76.83 & 77.02\\
\texttt{yeast} & 16.71 & \textbf{74.26} & 62.07 & 19.34 & 71.07 & 71.49\\
\hline
\end{tabular}
\end{table}

We can see that the best results are obtained with the non-enhanced CA
matrix with the average linkage strategy.

The CA matrix enhancement does not seem to have an impact on the single
linkage strategy, where the improvements are very limited. As the matrix
enhancement strategy improves negative pairs (i.e., when
\(\delta_{i,j} = 0\)), this may not impact strong spurious links which
impact negatively the single linkage strategy.

For maximum linkage, the enhancement has a positive impact (see
\texttt{abalone}, \texttt{landstat}, or \texttt{yeast}), where this
strategy becomes as good as the average linkage. In the maximum linkage
strategy, if only one pair is of low quality, it prevents two clusters
from merging. When enhancing the negative pairs, low quality pairs are
improved, helping merging similar clusters.

\begin{table}[h]
  \centering
  \caption{Density for hierarchical algorithms}
  \begin{tabular}{|c|c|c|c|c|c|c|}
    \hline
		Dataset & R-SL & R-AL & R-ML & E-SL & E-AL & E-ML\\
		\hline
\texttt{abalone} & 14.0 & \textbf{42.3} & 29.3 & 13.9 & 39.9 & 41.2\\
\texttt{advertisement} & 48.3 & \textbf{75.0} & 69.5 & 48.0 & 69.8 & 69.7\\
\texttt{breastcancer} & 59.1 & 77.3 & \textbf{77.8} & 51.9 & 59.3 & 63.5\\
\texttt{contraceptive} & 22.6 & \textbf{57.6} & 54.6 & 16.5 & 49.7 & 54.4\\
\texttt{ecoli} & 65.4 & 80.4 & \textbf{81.4} & 56.6 & 73.2 & 75.6\\
\texttt{glass} & 92.1 & 93.7 & \textbf{94.6} & 85.5 & 89.5 & 90.1\\
\texttt{ionosphere} & 66.8 & 85.1 & \textbf{87.5} & 69.9 & 78.3 & 82.5\\
\texttt{landstat} & 6.2 & \textbf{45.5} & 30.8 & 6.1 & 38.2 & 40.7\\
\texttt{user\_knowledge} & 53.4 & 76.9 & \textbf{77.34} & 47.70 & 67.00 & 72.6\\
\texttt{yeast} & 10.2 & \textbf{57.6} & 48.10 & 10.05 & 47.74 & 51.2\\
\hline
\end{tabular}
\end{table}

For density, non-enhanced results are one average better than the
enhanced ones. The disparities between algorithms are amplified, where
the difference between the highest and lowest values often exceeds more
than 10 points. Here, R-AL and R-ML perform best.

\hypertarget{comparison-1}{%
\subsection{Comparison}\label{comparison-1}}

The following tables record the best results of each group (hierarchical
\ref{hierarchical} and non-hierarchical \ref{non-hierarchical}). The
results obtained with the simple \(k\)-means (KM) and the bisecting
\(k\)-means (H-KM) are recorded here, plus the values for the max and
mean NMI obtained with zero effort.

\begin{table}[h]
  \centering
  \caption{NMI comparison}
  \begin{tabular}{|c|c|c|c|c|c|c|}
    \hline
Dataset & Max & Mean & KM & H-KM & Best-NH & Best-H\\
\hline

\texttt{abalone} & 85.34 & 84.23 & 80.48 & \textbf{82.35} & 81.02 & 81.31\\
\texttt{advertisement} & 81.54 & 77.16 & 81.82 & 80.29 & 80.61 & \textbf{81.93}\\
\texttt{breastcancer} & 74.96 & 69.94 & 73.41 & 67.12 & 74.87 & \textbf{77.76}\\
\texttt{contraceptive} & 72.57 & 69.31 & 75.07 & 74.48 & 74.39 & \textbf{75.28}\\
\texttt{drybean} & 82.67 & 79.73 & 80.09 & \textbf{80.77} & 78.90 & NA\\
\texttt{ecoli} & 80.67 & 77.62 & 80.65 & 78.21 & 80.91 & \textbf{82.61}\\
\texttt{gisette} & 58.66 & 56.1 & \textbf{60.81} & 58.34 & 59.82 & NA\\
\texttt{glass} & 85.96 & 82.35 & 77.24 & 74.34 & \textbf{84.3} & 84.04\\
\texttt{ionosphere} & 74.08 & 68.94 & 72.15 & 68.28 & 74.14 & \textbf{76.47}\\
\texttt{landstat} & 78.01 & 75.27 & \textbf{77.2} & 76.67 & 76.06 & 76.26\\
\texttt{musk} & 82.06 & 77.83 & \textbf{79.51} & 79.29 & 78.78 & NA\\
\texttt{nursery} & 73.89 & 67.9 & 64.59 & \textbf{67.05} & 65.06 & NA\\
\texttt{pendigits} & 83.88 & 81.33 & 78.95 & 79.64 & \textbf{79.69} & NA\\
\texttt{user\_knowledge} & 74.32 & 69.88 & 76.63 & 75.21 & 76.33 & \textbf{77.89}\\
\texttt{yeast} & 71.86 & 69.12 & \textbf{74.77} & 72.42 & 72.83 & 74.26\\
\hline
\end{tabular}
\end{table}

We see that the NMI are comparable between the different algorithms,
where the difference between the highest and lowest score is relatively
small. If we count the number of times an algorithm is first,
hierarchical are first, despite the fact they have not been applied on
all datasets.

\begin{table}[h]
  \centering
  \caption{NMI comparison}
  \begin{tabular}{|c|c|c|c|c|c|c|}
    \hline
Dataset & Max & Mean & KM & H-KM & Best-NH & Best-H\\
\hline
\texttt{abalone} & 73.9 & 66.9 & \textbf{51.6} & 49.9 & 47.5 & 42.3\\
\texttt{advertisement} & 89.7 & 75.1 & \textbf{80.9} & 78.0 & 77.2 & 75.0\\
\texttt{breastcancer} & 81.7 & 69.7 & \textbf{85.7} & 79.9 & 82.3 & 77.8\\
\texttt{contraceptive} & 62.9 & 50.1 & \textbf{60.2} & 58.9 & 57.9 & 57.6\\
\texttt{drybean} & 84.6 & 65.8 & \textbf{55.4} & 54.9 & 52.1 & NA\\
\texttt{ecoli} & 76.5 & 69.4 & \textbf{83.3} & 81.2 & 81.2 & 81.4\\
\texttt{gisette} & 45.1 & 38.4 & \textbf{35.6} & 33.2 & 34.5 & NA\\
\texttt{glass} & 88.9 & 81.1 & \textbf{96.7} & 91.7 & 94.6 & 94.6\\
\texttt{ionosphere} & 83.1 & 72.6 & \textbf{88.9} & 81.7 & 86.3 & 87.5\\
\texttt{landstat} & 72.0 & 60.7 & \textbf{56.9} & 55.8 & 51.6 & 45.5\\
\texttt{musk} & 85.0 & 66.9 & \textbf{55.1} & 55.0 & 51.8 & NA\\
\texttt{nursery} & 67.4 & 47.1 & 29.8 & \textbf{32.3} & 30.8 & NA\\
\texttt{pendigits} & 78.1 & 66.3 & \textbf{46.7} & 45.8 & 43.1 & NA\\
\texttt{user\_knowledge} & 74.2 & 63.1 & \textbf{79.8} & 78.3 & 78.0 & 77.3\\
\texttt{yeast} & 63.0 & 55.4 & \textbf{61.5} & 59.1 & 58.1 & 57.6\\
\hline
\end{tabular}
\end{table}

In terms of density, unsurprisingly, the simple \(k\)-means outperforms,
and the second place is often reached by the bisecting \(k\)-means.
While there was large discrepancy within an algorithm category in the
previous density tables, here, the difference between the min and the
max is relatively small, unless for very few datasets (\texttt{musk} or
\texttt{nursery}) where this difference exceeds 10 points.

\clearpage

\pagebreak

\hypertarget{discussion}{%
\section{Discussion}\label{discussion}}

\hypertarget{is-nmi-a-pertinent-metrics}{%
\subsection{Is NMI a Pertinent
Metrics?}\label{is-nmi-a-pertinent-metrics}}

NMI is widely used in the literature about ensemble clustering.
Intuitively, the NMI is a good candidate to compare \emph{two}
clustering, as it measures the overlap between their clusters. However,
for ensemble clustering, we do not consider it as an ``optimal''
candidate for several reasons.

If we look at classification problems, the most common metric is
accuracy, where it ranges between \(0\) and \(1\). For a dataset with
\(k\) classes, the lowest accuracy we can obtain with zero effort is
\(\frac{1}{k}\), and the maximal reachable is \(1\), if data are fully
separable. With \(20\) classes, the accuracy should be above \(5\%\)
with a zero effort strategy (when selecting a label at random).

With the NMI, it is possible to reach something close to \(0\), if the
generated consensus is completely random. However, a simple still zero
effort strategy is to select one partitioning from the ensemble as the
final partitioning. In that case, the NMI is already large. Therefore,
the range of NMI is not \([0, 1]\), but a much narrow interval. We
cannot be happy with an NMI of \(75\%\) when the mean NMI of the
ensemble is already \(70\%\).

Next, given the experimental protocol, which is similar to many other
papers in the field, ensemble's partitionings have different cluster
numbers. Even if common boundaries perfectly overlap (e.g., if \(\pi_A\)
has \(k\) clusters and \(\pi_B\) has \(2k\) clusters, where each cluster
in \(\pi_A\) corresponds to exactly two clusters in \(\pi_B\)), the NMI
will never be equal to \(1\), whatever the number of clusters in
\(\pi^*\). Therefore, the upper bound can never be reached.

For these two reasons, we do not consider the NMI as a good metric.

\hypertarget{why-not-using-ari}{%
\subsubsection{Why not using ARI?}\label{why-not-using-ari}}

ARI suffers from the same issues as the NMI. NMI comes from the
information theory field, while ARI plays with possible permutations.
While the computation of each metric is different, their behaviors are
similar.

\hypertarget{is-there-a-better-metric}{%
\subsubsection{Is There a Better
Metric?}\label{is-there-a-better-metric}}

While these experimental results make us doubtful about the usefulness
of the NMI, we have no alternative solution to suggest yet. Density is
an option, but it does not pay attention to inter-density and
information loss.

Internal metrics such as silhouette are ill-suited to the ensemble
clustering problem because the consensus algorithms do not have access
to these initial features. The results of such an evaluation are more
likely to be random, depending on the data and selected clustering
algorithms.

\hypertarget{why-bisecting-k-means-under-perform-k-means}{%
\subsection{\texorpdfstring{Why Bisecting \(k\)-means Under-Perform
\(k\)-means?}{Why Bisecting k-means Under-Perform k-means?}}\label{why-bisecting-k-means-under-perform-k-means}}

The bisecting approach works by splitting recursively the current
cluster. When a split is done, the two resulting parts become
independent and there is no way to update them by exchanging elements.

At the start, when the number of elements is large, most of the items
will be well segregated. However, because the split is rough, items near
the boundary are likely to be assigned to the wrong cluster. Therefore,
these errors cannot be rectified and penalized the final density result.
Nevertheless, the differences are reasonable and do not discourage from
using the bisecting \(k\)-means.

\hypertarget{which-algorithm-should-be-used}{%
\subsection{Which Algorithm Should be
Used?}\label{which-algorithm-should-be-used}}

In any case, agglomerative algorithms are too costly to run on large
datasets, so this category is gently discarded from the choice.

Next, MCLA and SEC lead to good results in terms of NMI, as they provide
the best results in their category. Nevertheless, when we compare them
individually to the \(k\)-means, they are weaker than it with less
consistent results. Therefore, we would advise using the \(k\)-means
algorithm to identify an optimal partitioning with a large NMI.

If we compare the bisecting \(k\)-means to the \(k\)-means, they are
equivalent in terms of NMI. However, in terms of density, the
\(k\)-means leads to the best outcome.

The choice between the \(k\)-means or the bisecting version depends on
the application, where in the case where the number of clusters is
unknown, it is preferable in terms of runtime to use the bisecting
\(k\)-means, even if the density is not fully maximized.

\hypertarget{conclusion}{%
\section{Conclusion}\label{conclusion}}

In this article, we presented a way to compute efficiently the density
of a partitioning. Instead of using the square co-association matrix,
the density can be directly computed on the raw binary matrix \(H\),
leading to a non-negligible speed up, moving from a quadratic to a
linear cost. We proved that maximizing the partitioning density is
equivalent to minimizing the \(k\)-means loss. Therefore, the density
can be simply maximized by clustering the binary matrix \(H\) with the
simple \(k\)-means.

We compared the \(k\)-means approach to previous state-of-the-art
algorithms. Additionally, because the number of clusters might be
unknown, we compared the \(k\)-means to the bisecting \(k\)-means to
study how large the difference is.

The results obtained with the \(k\)-means in terms of NMI are comparable
to the best state-of-the-art consensus algorithms, to the difference of
its simplicity. We experimentally verified that the \(k\)-means
maximizes the density consistently across several datasets. When using
the bisecting \(k\)-means, the NMI is unaffected, while the density is a
little bit lowered. Therefore, it is a good alternative when necessary.

In this paper, we argued that the NMI is not the best metric we could
have in this field, as the range of variation is limited. The density we
defined is not the best alternative either, as it only considers
intra-similarity and does not pay attention to relevant relationships
which have been discarded. Additional research needs to be performed in
this direction for providing a better way to compare strength and
weaknesses of consensus algorithms.

\clearpage

\pagebreak

\bibliographystyle{unsrtnat}

\bibliography{references}  

\end{document}